\def\year{2021}\relax
\documentclass[letterpaper]{article} 
\usepackage{aaai21}  
\usepackage{times}  
\usepackage{helvet} 
\usepackage{courier}  
\usepackage[hyphens]{url}  
\usepackage{graphicx} 
\usepackage{amsmath}
\usepackage{natbib}  
\usepackage{caption} 
\usepackage[switch]{lineno}
\usepackage{multirow}

\title{Dual-Level Collaborative Transformer for Image Captioning}

\author {
    Yunpeng Luo\textsuperscript{\rm 1},
    Jiayi Ji\textsuperscript{\rm 1},
    Xiaoshuai Sun\textsuperscript{\rm 1}\thanks{Corresponding Author},
    Liujuan Cao\textsuperscript{\rm 1}, \\
    Yongjian Wu\textsuperscript{\rm 3},
    Feiyue Huang\textsuperscript{\rm 3},
    Chia-Wen Lin\textsuperscript{\rm 4},
    Rongrong Ji\textsuperscript{\rm 1,2} \\
}

\affiliations {
    \textsuperscript{\rm 1} Media Analytics and Computing Lab, Department of Artificial Intelligence, \\ School of Informatics, Xiamen University, 361005, China \\
    \textsuperscript{\rm 2} Institute of Artificial Intelligence, Xiamen University \\
    \textsuperscript{\rm 3} Tencent Youtu Lab \qquad\qquad\  
    \textsuperscript{\rm 4} National Tsing Hua University \\
    lyricpoem1997@gmail.com, jjyxmu@gmail.com,xssun@xmu.edu.cn, caoliujuan@xmu.edu.cn, \\ littlekenwu@tencent.com, garyhuang@tencent.com, cwlin@ee.nthu.edu.tw, rrji@xmu.edu.cn
}

\begin{document}
\maketitle

\begin{abstract}
Descriptive region features extracted by object detection networks have played an important role in the recent advancements of image captioning.
However, they are still criticized for the lack of contextual information and fine-grained details, which in contrast are the merits of traditional grid features.
In this paper, we introduce a novel Dual-Level Collaborative Transformer (DLCT) network to realize the complementary advantages of the two features.
Concretely, in DLCT, these two features are first processed by a novel \emph{Dual-way Self Attenion} (DWSA) to mine their intrinsic properties, where a \emph{Comprehensive Relation Attention} component is also introduced to embed the geometric information.
In addition, we propose a \emph{Locality-Constrained Cross Attention} module to address the semantic noises caused by the direct fusion of these two features, where a geometric alignment graph is constructed to accurately align and reinforce region and grid features.
To validate our model, we conduct extensive experiments on the highly competitive MS-COCO dataset, and achieve new state-of-the-art performance on both local and online test sets, \emph{i.e.}, 133.8\% CIDEr on \emph{Karpathy} split and 135.4\% CIDEr on the official split. Code is available at https://github.com/luo3300612/image-captioning-DLCT.
\end{abstract}

\section{Introduction}
Image captioning is the task of generating a descriptive statement automatically for an input image. 
Its main challenges not only lie in the comprehensive understanding of objects and relationships in the image, but also in the generation of fluent sentences that match the visual semantics.
With years of developments, the great success of image captioning has been supported by a flurry of methods \cite{rennie2017self,anderson2018bottom,Zhou2020UnifiedVP} and benchmark datasets \cite{lin2014microsoft}.

Among these advancements, a milestone in image captioning is the introduction of visual region features extracted by object detection networks~\cite{anderson2018bottom}, \emph{e.g}., Faster R-CNN~\cite{ren2015faster}. 
Compared with the grid features\footnote{The feature maps of the pre-trained convolution neural networks (CNN).} used in earlier methods \cite{vinyals2015show}, region features can provide object-level information, since most salient regions in an image can be recognized and represented by a feature vector. Hence, region features
greatly reduce the difficulty of visual-semantic embeddings, based on which recent endeavors have greatly boosted the performance of image captioning \cite{huang2019attention, cornia2020meshed, pan2020x}.

\begin{figure}
    \centering
    \includegraphics[width=0.45\textwidth]{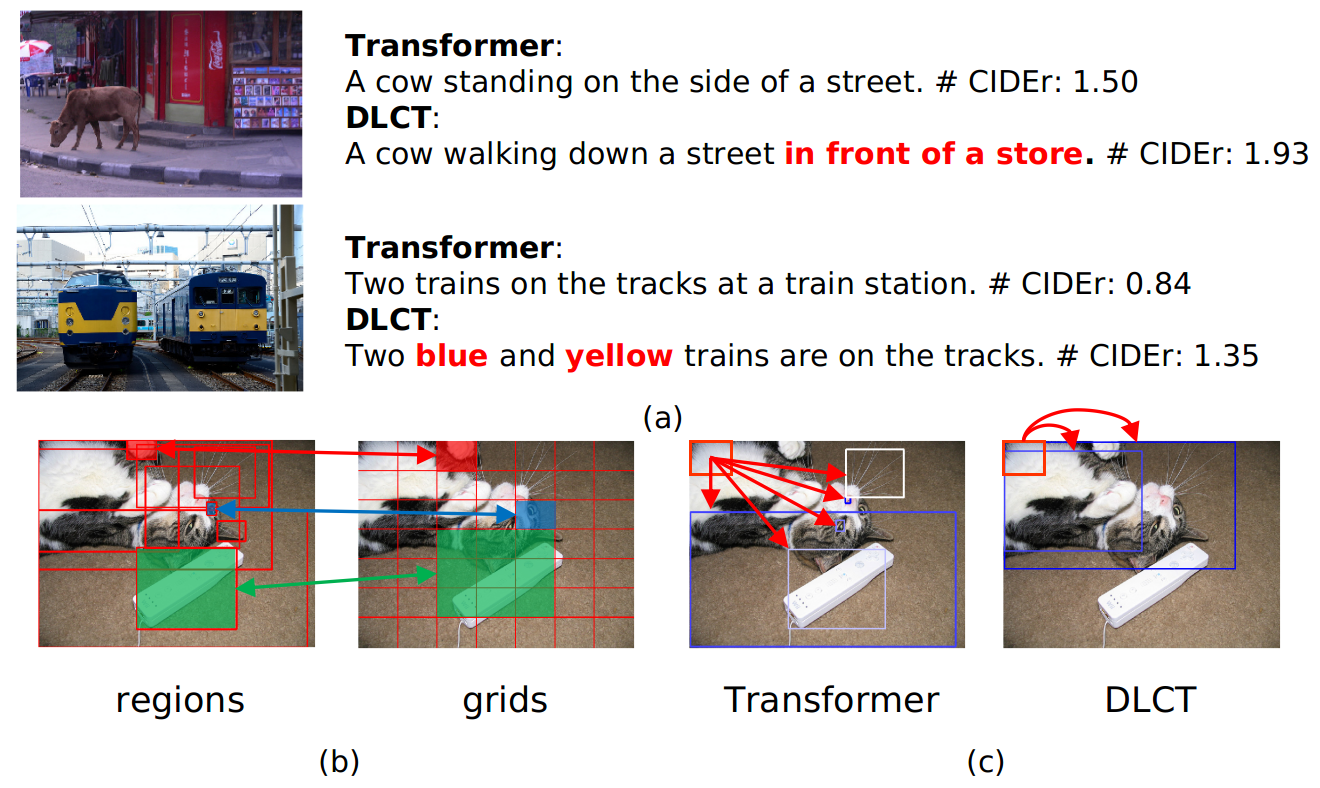}
\caption{(a) Limitations of region features on characterizing contextual (up) and detailed (down) information.
(b) An example of region features (left), grid features (right) and their geometric alignment. Our model enables the interaction between two kinds of features based on their semantic alignment constructed according to their geometric properties.
(c) An illustration of \textbf{semantic noise problem}. Blue regions are the top-k attended regions (from deep to shallow) by the red grid.
In Transformer (left), the top-5 attended regions are all semantically unrelated. In our DLCT (right), the red grid only attends to two semantically related regions. 
}
\label{fig1}      
\end{figure}

Despite the great success, region features are still criticized for the lack of contextual information and fine-grained details.
As illustrated in Fig.1-(a), the detected regions may not cover the entire image, leading to the inability to correctly describe the global scenes, \emph{e.g.}, \textit{in front of a store}. Meanwhile, each region is represented by a single feature vector, which inevitably loses object details in large amounts, \emph{e.g.}, the colors of trains.
However, these shortcomings are the merits of grid features which in contrast cover all the content of a given image in a more fragmented form.

To this end, it is a natural thought to use both features as the visual input, which however results in a new issue.
To explain, most recent methods in image captioning \cite{huang2019attention, cornia2020meshed, pan2020x} use the \emph{self-attention} modules to model the relationships of visual features.
Under this setting, the direct use of two sources of features is prone to producing semantic noises during the attention process.
For instance, a grid may interact with incorrect regions just because they have similar appearances, \emph{e.g.}, the cat's belly and the white remote controller, as shown in Fig.\ref{fig1}-(c).
Such a case not only hinders the complementarity of two features but also degrades the overall performance, \emph{i.e.}, using two features might be worse than using one, which has also been validated in Tab.\ref{table5}.

\begin{figure*}
    \centering
    \includegraphics[width=0.95\textwidth]{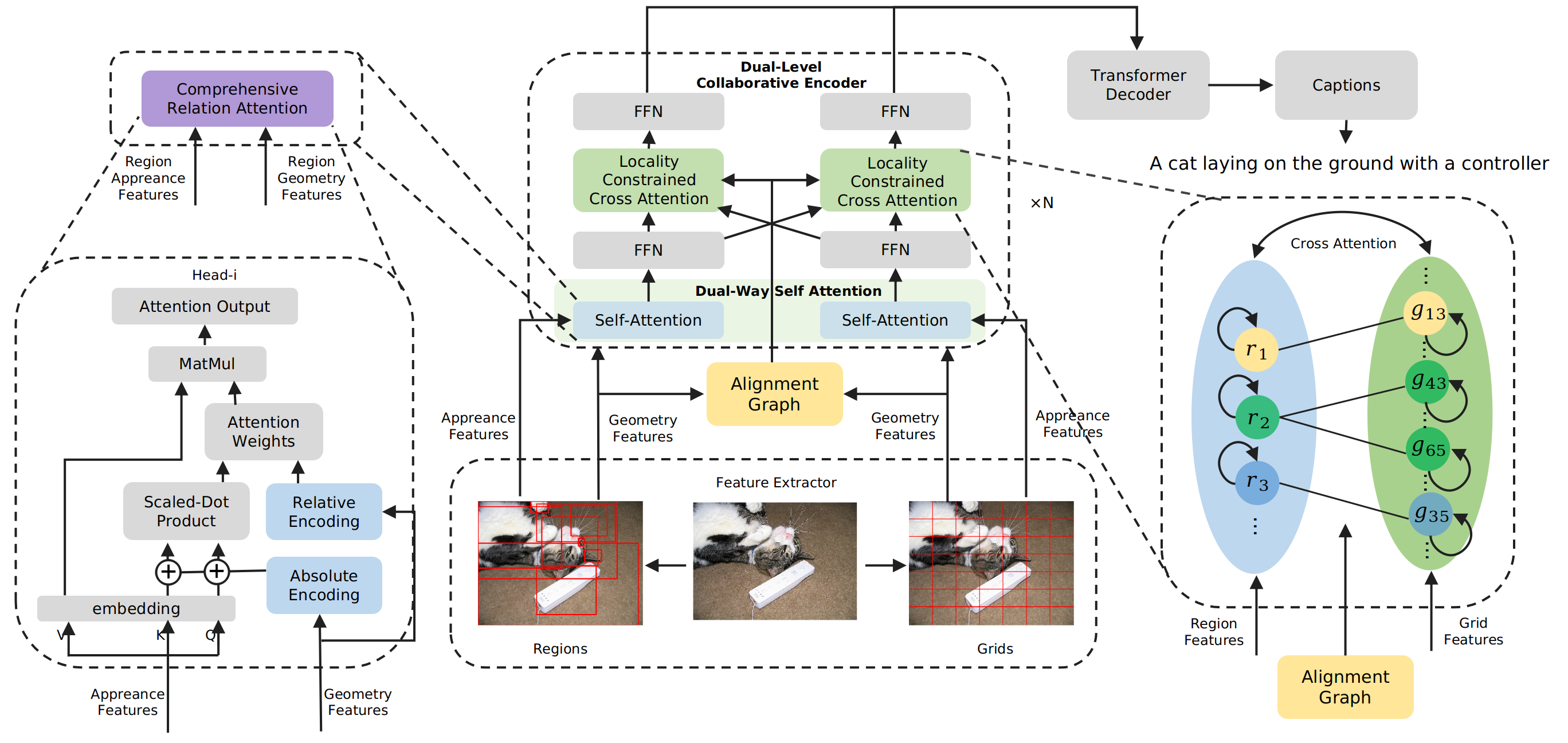}
    \caption{Overview of the proposed Dual-Level Collaborative Transformer architecture.
We devise the Comprehensive Relation Attention to integrate position information in both absolute and relative manners.
The Dual-Way Self Attention is applied to mine the intrinsic properties of two kinds of features,
followed by the Locality-Constrained Cross Attention (LCCA) which enables the interaction between regions and grids.
With the geometric alignment graph, LCCA can eliminate semantic noises and achieve inter-level fusion effectively.}
    \label{fig2}       
\end{figure*}

In this paper, we propose a novel \emph{Dual-Level Collaborative Transformer} (DLCT) network to realize the complementary advantages of region and grid features for image captioning.
Concretely, as shown in Fig.\ref{fig2}, the two sources of features are first processed by a novel \emph{Dual-Way Self-Attention} (DWSA) module to explore their intrinsic properties, where a \emph{Comprehensive Relation Attention} (CRA) scheme is equipped to embed absolute and relative geometry information of input features.
In addition, we further propose a \emph{Locality-Constrained Cross Attention} (LCCA) module to address the aforementioned degradation issue, where a geometric alignment graph is constructed to guide the semantic alignment between two sources of features.
With this geometric alignment graph, LCCA can accurately enable the interaction between features of two sources. More importantly, it can reinforce each type of feature by cross-attention fusions, such as transferring objectness information from region features to grid ones and supplementing fine-grained details from grid features to region ones.

To validate the proposed DLCT, we conduct extensive experiments on MS-COCO dataset \cite{lin2014microsoft}, and achieve new state-of-the-art performances for image captioning, \emph{i.e.}, 133.8\% CIDEr scores on \emph{Karpathy} test set \cite{karpathy2015deep} and 135.4\% CIDEr scores on the online test.

We summarize the contributions of this paper as follows:
\begin{itemize}
\item We propose an \emph{Dual-level Collaborative Transformer} network to achieve the complementarity of region and grid features. Extensive experiments on MS-COCO dataset demonstrate the superior performance of our method compared with the state-of-the-arts.
\item We propose Locality-Constrained Cross Attention to address the issue of semantic noise aroused by the direct fusion of two sources of features. With the constructed geometric alignment graphs, LCCA can not only enables the interaction between features of different sources accurately, but also reinforce each kind of feature via cross-attention fusions.
\item To our best knowledge, we also present the first attempt to explore the absolute position information for image captioning. By integrating absolute and relative location information, we further improve the modeling of intra- and inter-level relationships.
\end{itemize}

\section{Related Work}
Existing image captioning approaches typically follow the encoder-decoder architecture \cite{xu2015show,huang2019attention, guo2020normalized,cornia2020meshed,Zhao2020MemCapMS,Seo2020ReinforcingAI}, which takes an image as input and generates a description in the form of natural language. Earlier works \cite{xu2015show,lu2017knowing,jiang2020defense} apply grid-based features as input to generate captions, which are fixed-size patches extracted from the CNN \cite{he2016deep,Lin_2020_CVPR} model. Recently, region-level features extracted by Faster-RCNN \cite{ren2015faster} have also been introduced to captioning models, significantly improving the quantitative performance of image captioning \cite{anderson2018bottom,herdade2019image,huang2019attention,cornia2020meshed,guo2020normalized}. Nevertheless, they have a deficiency of predicting sentences by using only one kind of feature.

HAN \cite{wang2019hierarchical} proposes a hierarchical attention network to combine text, grids, and regions with a relation module to exploit the inherent relationship among diverse features. However, it fails to integrate location information of visual features and coarsely model appearance relationship while ignoring to filter semantic noises. GCN-LSTM \cite{yao2018exploring} and Object Relation Transformer \cite{herdade2019image} utilize bounding boxes of regions to model location relationships between regions in a relative manner. However, by modeling location relatively, they can integrate appearance features and geometry features but still fail to grab the absolute locations of features in an image.

\section{Dual-Level Collaborative Transformer}
In this section, we introduce a novel image captioning model, named Dual-Level Collaborative Transformer, which uses both grid and region features to achieve the complementarity of them. 
The overall structure of our model is illustrated in Fig. \ref{fig2}.

\subsection{Integrating Position Information}
Previous methods only model location relationships of regions in a relative manner.
Thus we propose Comprehensive Relation Attention (CRA) to model complex visual and location relationships between input features by integrating both absolute and relative location information.

\subsubsection{Absolute Postional Encoding}
Absolute positional encoding (APE) tells the model where the feature is, which is important information. Suppose there are two objects with identical appearance features: one locates in the corner and the other locates at the center. In this case, APE facilitates the model to distinguish them accurately.
For APE, we consider two kinds of visual features, \emph{i.e.}, grids and regions. For grids, we use the concatenation of two 1-d sine and cosine embeddings to get the grid positional encoding (GPE):
\begin{equation}
    GPE(i,j)=[PE_i;PE_j],
\end{equation}
where $i$,$j$ are the row index and column index of the grid and $PE_i,PE_j\in \mathbf{R}^{d_{model}/2}$ are defined as:
\begin{equation}
    \begin{aligned}
        &PE(pos,2k)=\sin(pos/10000^{2k/(d_{model}/2)}),\\
        &PE(pos,2k+1)=\cos(pos/10000^{2k/(d_{model}/2)}),  
    \end{aligned}
\end{equation}
where $pos$ denotes the position and $k$ is the dimension. For regions, we embed 4-d bounding box $B_i=(x_{min},y_{min},x_{max},y_{max})$ in region positional encoding (RPE):
\begin{equation}
    RPE(i)=B_iW_{emb},
\end{equation}
where $i$ is the index of box, $(x_{min},y_{min})$ and $(x_{max}, y_{max})$ respectively denote the top-left and bottom-right corners of the box and $W_{emb} \in \mathbf{R}^{d_{model}\times 4}$ is an embedding parameter matrix.

\subsubsection{Relative Positional Encoding}
To better integrate relative location information of visual features, we add relative location information according to the geometric structure of bounding boxes.
The bounding box of a region can be represented as $(x, y, w, h)$ where $x$, $y$, $w$, and $h$ denote the box’s center coordinates and its width and height.
Note that a grid is a special case of a bounding box.
So grids can also be represented as $(x, y, w, h)$ according to its respective field.
Thus for $box_i$ and $box_j$, we can represent their geometric relationship as a 4-d vector:
\begin{equation}
    \begin{aligned}
        &\Omega(i,j) = \\
        &\left( \log{(\frac{\left|x_i-x_j\right|}{w_i})},\log{(\frac{\left|y_i-y_j\right|}{h_i})},\log{(\frac{w_i}{w_j})},\log{(\frac{h_i}{h_j})} \right)^T.
    \end{aligned}
\end{equation}
Then $\Omega(i,j)$ is embeded in a high-dimensional embedding by the Emb method in \cite{vaswani2017attention}.
Finally, $\Omega(i,j)$ is mapped to a scalar which conveys the geometric relationship between two boxes:
\begin{equation}
    \Omega(i,j) = \mbox{ReLU}(\mbox{Emb}(\Omega(i,j))W_G),
\end{equation}
where $W_G$ is a learned parameter matrix.

\subsubsection{Comprehensive Relation Attention}
Once absolute information and relative information are extracted, we can integrate them by Comprehensive Relation Attention (CRA).
For APE, we modify the queries and keys at the attention layer:
\begin{equation}
    W=\frac{(Q+pos_q)(K+pos_k)^T}{\sqrt{d_k}},
\end{equation}
where $pos_q$ and $pos_k$ are APE of queries and keys respectively.
Then we utilize relative location information to adjust attention weights by:
\begin{equation}
    W'_{ij} = W_{ij} + \log(\Omega(i,j)).
\end{equation}
Finally, softmax is applied to normalize weights and calculate the outputs of CRA.
Our Multi-Head CRA (MHCRA) can be formalized as:
\begin{equation}
    \mbox{MHCRA}(Q,K,V)= \mbox{Concat}(\mbox{head}_1,\cdots,\mbox{head}_h)W^O,
\end{equation}
\begin{equation}
    \mbox{head}_i=\mbox{CRA}(QW_i^Q,KW_i^K,VW_i^V,pos_q,pos_k,\Omega),
\end{equation}
where
\begin{equation}
    \begin{aligned}
        &\mbox{CRA}(Q,K,V,pos_q,pos_k,\Omega)=\\
        &\mbox{softmax}(\frac{(Q+pos_q)(K+pos_k)^T}{\sqrt{d_k}}+ \log(\Omega))V.
    \end{aligned}
\end{equation}

\subsection{Dual-Level Collaborative Encoder}
Given an image, we firstly extract its grid and region features respectively dubbed as $V_G=\{v_i\}^{N_G}$ and $V_R=\{v_i\}^{N_R}$.
$N_G$ and $N_R$ are numbers of corresponding features.
Our encoder consists of two sub-modules: Dual-Way Self Attention and Locality-Constrained Cross Attention.

\subsubsection{Dual-Way Self Attention}
In general, visual features are extracted by locally-connected convolutions, which make them isolated and relation-agnostic.
It is believed that Transformer Encoder contributes significantly to the performance of image captioning,
because it can model relationships between the inputs to enrich visual features by self-attention.
To better model intra-level relationships of two kinds of features, we devise a Dual-Way Self Attention (DWSA) which consists of two independent self-attention modules.

Specifically, the hidden states of regions $\mathbf{H}^{(l)}_r$ and grids $\mathbf{H}^{(l)}_g$ are fed into the $(l+1)$-th DWSA to learn relation-aware representaion:
\begin{equation}
    \mathbf{C}^{(l)}_r = \mbox{MHCRA}(\mathbf{H}^{(l)}_r,\mathbf{H}^{(l)}_r,\mathbf{H}^{(l)}_r,\mbox{RPE},\mbox{RPE},\Omega_{rr}),
\end{equation}
\begin{equation}
    \mathbf{C}^{(l)}_g = \mbox{MHCRA}(\mathbf{H}^{(l)}_g,\mathbf{H}^{(l)}_g,\mathbf{H}^{(l)}_g,\mbox{GPE},\mbox{GPE},\Omega_{gg}).
\end{equation}
where $\mathbf{H}_r^{(0)}=V_R$, $\mathbf{H}_g^{(0)}=V_G$.
$\Omega^{rr}$ and $\Omega^{gg}$ are relative location matrix of regions and grids respectively.
Then we adopt two independent position-wise feedforward networks FFN for each type of visual features:
\begin{equation}
    \mathbf{C}_r^{'(l)}=\mbox{FFN}_r(\mathbf{C}^{(l)}_r),
\end{equation}
\begin{equation}
    \mathbf{C}_g^{'(l)}=\mbox{FFN}_g(\mathbf{C}^{(l)}_g).
\end{equation}
After that, the relation-aware representations are fed into the next module.

\begin{figure}
    \centering
    \includegraphics[width=0.45\textwidth]{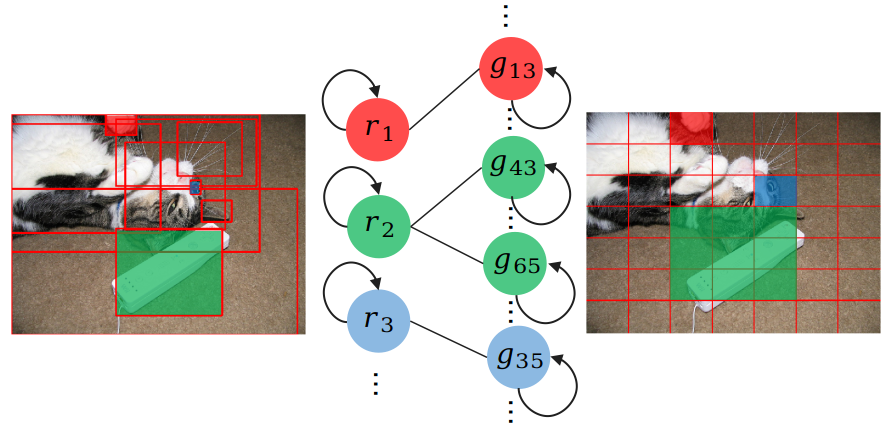}
    \caption{Example of a geometric alignment graph. Regions and grids with intersections (highlighted with the same color) are connected by undirected edges to eliminate semantically unrelated information. Note each node has a self-connected edge.}
    \label{fig3}       
\end{figure}

\subsubsection{Locality-Constrained Cross Attention}
We propose Locality-Constrained Cross Attention (LCCA) to model complex interactions between regions and grids for inter-level fusion.
To avoid introducing semantic noises, we first create a geometric alignment graph $G=(V, E)$.
All region and grid features are represented as independent nodes to form a visual node set $V$.
For edge set $E$, a grid node is connected to a region node if and only if their bounding boxes have intersections.
Following the above rules, we can construct an undirected graph, as illustrated in Fig. \ref{fig3}.
Based on the geometric alignment graph, we apply LCCA to identify attention across two different kinds of visual feature fields: the source field and the target field.
In LCCA, the source field serves as queries and the target field serves as keys and values.
LCCA aims at reinforcing representation of the source field by embedding information of the target field into the source field.
Like Equ. (1)(2), we integrate the absolute and relative location information to get the weight matrix $W^{'}$ and normalize it:
\begin{equation}
    \alpha_{ij}=\frac{e^{W^{'}_{ij}}}{\sum_{j\in A(v_i)}e^{W^{'}_{ij}}},
    \label{eq1}
\end{equation}
where $v_i$ is the visual node and $A(v_i)$ is the set of neighboring visual nodes of $v_i$.
The weighted sum is applied as
\begin{equation}
    \mathbf{M}_i = \sum_{j\in A(v_i)}\alpha_{ij}^{(l)}V_j,
    \label{eq2}
\end{equation}
where $V_j$ is the $j$-th visual node value. For simplicity, we formulate this stage as
\begin{equation}
    \mathbf{M}=\mathop{\mbox{graph-softmax}}\limits_G(W')V,
\end{equation}
where graph-softmax assign 0 weight to non-neighboring visual nodes and apply softmax like Equ. (\ref{eq1}) based on $G$.
Overall, our Multi-Head LCCA (MHLCCA) can be formulated as
\begin{equation}
    \mbox{MHLCCA}(Q,K,V)= \mbox{Concat}(\mbox{head}_1,\cdots,\mbox{head}_h)W^O,
\end{equation}
\begin{equation}
    \mbox{head}_i=\mbox{LCCA}(QW_i^Q,KW_i^K,VW_i^V,pos_q,pos_k,\Omega,G),
\end{equation}
where
\begin{equation}
    \begin{aligned}
        &\mbox{LCCA}(Q,K,V,pos_q,pos_k,\Omega,G)=\\
        &\mathop{\mbox{graph-softmax}}\limits_G(\frac{(Q+pos_q)(K+pos_k)^T}{\sqrt{d_k}}+ \log(\Omega))V.
    \end{aligned}
\end{equation}
In this stage, the grid features and region features serve as the source field and target field alternately.
For the $l$-th output of DWSA:
\begin{equation}
    \begin{aligned}
        &\mathbf{M}_r^{(l)}=\mbox{MHLCCA}(\mathbf{C}_r^{'(l)},\mathbf{C}_g^{'(l)},\mathbf{C}_g^{'(l)},\mbox{RPE},\mbox{GPE},\Omega_{rg},G),
    \end{aligned}
\end{equation}
\begin{equation}
    \begin{aligned}
        &\mathbf{M}_g^{(l)}=\mbox{MHLCCA}(\mathbf{C}_g^{'(l)},\mathbf{C}_r^{'(l)},\mathbf{C}_r^{'(l)},\mbox{GPE},\mbox{RPE},\Omega_{gr},G),
    \end{aligned}
\end{equation}
where $\Omega_{rg}$ is the relative position matrix between regions and grids and $\Omega_{gr}$ is the relative position matrix between grids and regions.

By LCCA, we embed regions into grids and vise versa to reinforce two kinds of features.
Specifically, grid features attend to regions to get high-level object information, while regions attend to grids to supplement detailed and contextual information.
With the geometric alignment graph, LCCA constrains information from semantically unrelated visual features to eliminate semantic noises and apply cross-attention effectively.

Note that a region can align with one or more grids while a grid can align with zero or more regions.
There might exist a grid that aligns with no region. So we create a self-connected edge for each node in the geometric alignment graph.
Besides, self-connected edges give the attention module an extra choice of not attending to any other features.
In the $l$-th layer, the attention module is followed by two independent FFN like in DWSA:
\begin{equation}
    \mathbf{H}_r^{(l+1)}=\mbox{FFN}'_r(\mathbf{M}^{(l)}_r),
\end{equation}
\begin{equation}
    \mathbf{H}_g^{(l+1)}=\mbox{FFN}'_g(\mathbf{M}^{(l)}_g).
\end{equation}
Note that the output of LCCA serves as the input of DWSA. After multi-layer encoding, grid features and region features are concatenated and fed into decoder layers.

\begin{table*}[t]
	\centering
	\small
	\resizebox{0.95\textwidth}{!}{ 
    \begin{tabular}{l|c|c|c|c|c|c|cccccc}
        \hline
                                  & \multicolumn{6}{c|}{Single Model}                                                                                                                                      & \multicolumn{6}{c}{Ensemble Model}                                                                                                                                                                      \\ \hline
        Model                     & B-1                       & B-4                       & M                         & R                         & C                          & S                         & \multicolumn{1}{c|}{B-1}           & \multicolumn{1}{c|}{B-4}           & \multicolumn{1}{c|}{M}             & \multicolumn{1}{c|}{R}             & \multicolumn{1}{c|}{C}              & S             \\ \hline
        SCST (ResNet-101) \tiny{cvpr2017}        & -                         & 34.2                      & 26.7                      & 57.7                      & 114.0                      & -                         & \multicolumn{1}{c|}{-}             & \multicolumn{1}{c|}{35.4}          & \multicolumn{1}{c|}{27.1}          & \multicolumn{1}{c|}{56.6}          & \multicolumn{1}{c|}{117.5}          & -             \\
        Up-Down (ResNet-101) \tiny{cvpr2018}     & 79.8                      & 36.3                      & 27.7                      & 56.9                      & 120.1                      & 21.4                      & \multicolumn{1}{c|}{-}             & \multicolumn{1}{c|}{-}             & \multicolumn{1}{c|}{-}             & \multicolumn{1}{c|}{-}             & \multicolumn{1}{c|}{-}              & -             \\
        HAN (ResNet-101)  \tiny{aaai2019}        & 80.9                      & 37.6                      & 27.8                      & 58.1                      & 121.7                      & 21.5                      & \multicolumn{1}{c|}{-}             & \multicolumn{1}{c|}{-}             & \multicolumn{1}{c|}{-}             & \multicolumn{1}{c|}{-}             & \multicolumn{1}{c|}{-}              & -             \\
        GCN-LSTM (ResNet-101) \tiny{eccv2018}    & 80.5                      & 38.2                      & 28.5                      & 58.5                      & 128.3                      & 22.0                      & \multicolumn{1}{c|}{80.9}          & \multicolumn{1}{c|}{38.3}          & \multicolumn{1}{c|}{28.6}          & \multicolumn{1}{c|}{58.5}          & \multicolumn{1}{c|}{128.7}          & 22.1          \\
        SGAE (ResNet-101) \tiny{cvpr2019}        & 80.8                      & 38.4                      & 28.4                      & 58.6                      & 127.8                      & 22.1                      & \multicolumn{1}{c|}{81.1}          & \multicolumn{1}{c|}{39.0}          & \multicolumn{1}{c|}{28.4}          & \multicolumn{1}{c|}{58.9}          & \multicolumn{1}{c|}{129.1}          & 22.2          \\
        ORT (ResNet-101)  \tiny{nips2019}        & 80.5                      & 38.6                      & 28.7                      & 58.4                      & 127.8                      & 22.1                      & \multicolumn{1}{c|}{-}    & \multicolumn{1}{c|}{-}    & \multicolumn{1}{c|}{-}             & \multicolumn{1}{c|}{-}    & \multicolumn{1}{c|}{-}    & -             \\
        SRT (ResNet-101)  \tiny{aaai2020}        & 80.3                      & 38.5                      & 28.7                      & 58.4                      & 129.1                      & 22.4                      & \multicolumn{1}{c|}{-}    & \multicolumn{1}{c|}{-}    & \multicolumn{1}{c|}{-}             & \multicolumn{1}{c|}{-}    & \multicolumn{1}{c|}{-}    & -             \\
        AoA (ResNet-101)  \tiny{iccv2019}        & 80.2                      & 38.9                      & 29.2                      & 58.8                      & 129.8                      & 22.4                      & \multicolumn{1}{c|}{81.6}          & \multicolumn{1}{c|}{40.2}          & \multicolumn{1}{c|}{29.3}          & \multicolumn{1}{c|}{59.4}          & \multicolumn{1}{c|}{132.0}          & 22.8          \\
        AoA (ResNeXt-101 Grid)  \tiny{iccv2019}        & 80.7                      & 39.0                      & 28.9                      & 58.7                      & 129.5                      & 22.6                      & \multicolumn{1}{c|}{-}          & \multicolumn{1}{c|}{-}          & \multicolumn{1}{c|}{-}          & \multicolumn{1}{c|}{-}          & \multicolumn{1}{c|}{-}          & -          \\
        HIP (SENet-154)  \tiny{iccv2019}        & -                      & 39.1                      & 28.9                      & 59.2                      & 130.6                      & 22.3                      & \multicolumn{1}{c|}{-}          & \multicolumn{1}{c|}{-}          & \multicolumn{1}{c|}{-}          & \multicolumn{1}{c|}{-}          & \multicolumn{1}{c|}{-}          & -          \\
        M2 (ResNet-101)   \tiny{cvpr2020}        & 80.8                      & 39.1                      & 29.2                      & 58.6                      & 131.2                      & 22.6                      & \multicolumn{1}{c|}{82.0}          & \multicolumn{1}{c|}{40.5}          & \multicolumn{1}{c|}{29.7}          & \multicolumn{1}{c|}{59.5}          & \multicolumn{1}{c|}{134.5}          & 23.5          \\
        M2 (ResNeXt-101 Region) \tiny{cvpr2020}  & \multicolumn{1}{l|}{80.6} & \multicolumn{1}{l|}{38.8} & \multicolumn{1}{l|}{29.0} & \multicolumn{1}{l|}{58.4} & \multicolumn{1}{l|}{130.8} & \multicolumn{1}{l|}{22.4} & \multicolumn{1}{c|}{-}             & \multicolumn{1}{c|}{-}             & \multicolumn{1}{c|}{-}             & \multicolumn{1}{c|}{-}             & \multicolumn{1}{c|}{-}              & -             \\
        M2 (ResNeXt-101 Grid) \tiny{cvpr2020}    & \multicolumn{1}{l|}{80.8} & \multicolumn{1}{l|}{38.9} & \multicolumn{1}{l|}{29.1} & \multicolumn{1}{l|}{58.5} & \multicolumn{1}{l|}{131.7} & \multicolumn{1}{l|}{22.6} & \multicolumn{1}{c|}{-}             & \multicolumn{1}{c|}{-}             & \multicolumn{1}{c|}{-}             & \multicolumn{1}{c|}{-}             & \multicolumn{1}{c|}{-}              & -             \\
        X-Transformer (ResNet-101) \tiny{cvpr2020} & 80.9                      & 39.7                      & \textbf{29.5}             & \textbf{59.1}             & 132.8                      & \textbf{23.4}             & \multicolumn{1}{c|}{81.7}          & \multicolumn{1}{c|}{40.7}          & \multicolumn{1}{c|}{\textbf{29.9}} & \multicolumn{1}{c|}{59.7} & \multicolumn{1}{c|}{135.3}          & \textbf{23.8} \\ 
        X-Transformer (ResNeXt-101 Grid) \tiny{cvpr2020} & 81.0                       & 39.7                      & 29.4             & 58.9             & 132.5                      & 23.1             & \multicolumn{1}{c|}{-}          & \multicolumn{1}{c|}{-}          & \multicolumn{1}{c|}{-} & \multicolumn{1}{c|}{-} & \multicolumn{1}{c|}{-}          & - \\ \hline
        Ours (ResNeXt-101)        & \textbf{81.4}             & \textbf{39.8}             & \textbf{29.5}             & \textbf{59.1}             & \textbf{133.8}             & 23.0                      & \multicolumn{1}{l|}{\textbf{82.2}} & \multicolumn{1}{l|}{\textbf{40.8}} & \multicolumn{1}{c|}{\textbf{29.9}}          & \multicolumn{1}{c|}{\textbf{59.8}} & \multicolumn{1}{c|}{\textbf{137.5}} & 23.3   \\ \hline      
        \end{tabular}
    }
    \caption{Performance comparisons on COCO Karpathy test split. B-1, B-4, M, R, C, and S are short for BLEU-1, BLEU-4, METEOR, ROUGE, CIDEr, SPICE scores, respectively. Note that 4 models are used for the ensemble. The backbone is listed in brackets.	 }\smallskip
    \label{table1}
\end{table*}

\begin{table*}[t]
	\centering
	\small
	\resizebox{0.95\textwidth}{!}{ 
    \begin{tabular}{lllllllllllllll}
        \hline
        \multicolumn{1}{c}{\multirow{2}{*}{Model}} & \multicolumn{2}{c}{B-1}                          & \multicolumn{2}{c}{B-2}                          & \multicolumn{2}{c}{B-3}                          & \multicolumn{2}{c}{B-4}                          & \multicolumn{2}{c}{M}                            & \multicolumn{2}{c}{R}                            & \multicolumn{2}{c}{C}                            \\
        \multicolumn{1}{c}{}                       & \multicolumn{1}{c}{c5} & \multicolumn{1}{c}{c40} & \multicolumn{1}{c}{c5} & \multicolumn{1}{c}{c40} & \multicolumn{1}{c}{c5} & \multicolumn{1}{c}{c40} & \multicolumn{1}{c}{c5} & \multicolumn{1}{c}{c40} & \multicolumn{1}{c}{c5} & \multicolumn{1}{c}{c40} & \multicolumn{1}{c}{c5} & \multicolumn{1}{c}{c40} & \multicolumn{1}{c}{c5} & \multicolumn{1}{c}{c40} \\ \hline
        SCST (ResNet-101)                                      & 78.1                   & 93.7                    & 61.9                   & 86.0                    & 47.0                   & 75.9                    & 35.2                   & 64.5                    & 27.0                   & 35.5                    & 56.3                   & 70.7                    & 114.7                  & 116.7                   \\
        Up-Down (ResNet-101)                                   & 80.2                   & 95.2                    & 64.1                   & 88.8                    & 49.1                   & 79.4                    & 36.9                   & 68.5                    & 27.6                   & 36.7                    & 57.1                   & 72.4                    & 117.9                  & 120.5                   \\
        HAN (ResNet-101)                                       & 80.4                   & 94.5                    & 63.8                   & 87.7                    & 48.8                      & 78.0                       & 36.5                   & 66.8                    & 27.4                   & 36.1                    & 57.3                   & 71.9                    & 115.2                  & 118.2                   \\
        GCN-LSTM (ResNet-101)                                   & 80.8                   & 95.2                    & 65.5                   & 89.3                    & 50.8                   & 80.3                    & 38.7                   & 69.7                    & 28.5                   & 37.6                    & 58.5                   & 73.4                    & 125.3                  & 126.5                   \\
        SGAE (ResNet-101)                                       & 81.0                   & 95.3                    & 65.6                   & 89.5                    & 50.7                   & 80.4                    & 38.5                   & 69.7                    & 28.2                   & 37.2                    & 58.6                   & 73.6                    & 123.8                  & 126.5                   \\
        AoA (ResNet-101)                                       & 81.0                   & 95.0                    & 65.8                   & 89.6                    & 51.4                   & 81.3                    & 39.4                   & 71.2                    & 29.1                   & 38.5                    & 58.9                   & 74.5                    & 126.9                  & 129.6                   \\
        HIP (SENet-154)                                       & 81.6                   & 95.9                    & 66.2                   & 90.4                    & 51.5                   & 81.6                    & 39.3                   & 71.0                    & 28.8                   & 38.1                    & 59.0                   & 74.1                    & 127.9                  & 130.2                   \\
        M2 (ResNet-101)                                        & 81.6                   & 96.0                    & 66.4                   & 90.8                    & 51.8                   & 82.7                    & 39.7                   & 72.8                    & 29.4                   & 39.0                    & 59.2                   & 74.8                    & 129.3                  & 132.1                   \\
        X-Transformer (ResNet-101)                              & 81.3                   & 95.4                    & 66.3                   & 90.0                    & 51.9                   & 81.7                    & 39.9                   & 71.8                    & 29.5                   & 39.0                    & 59.3                   & 74.9                    & 129.3                  & 131.4                   \\ 
        X-Transformer (SENet-154)                              & 81.9                   & 95.7                    & 66.9                   & 90.5                    & 52.4                   & 82.5                    & 40.3                   & 72.4                    & 29.6                   & 39.2                    & 59.5                   & 75.0                    & 131.1                  & 133.5       \\ \hline             
        DLCT (ResNeXt-101)                                      & 82.0          & 96.2           & 66.9          & 91.0           & 52.3          & 83.0           & 40.2          & 73.2           & 29.5          & 39.1           & 59.4          & 74.8           & 131.0         & 133.4        \\           
        DLCT (ResNeXt-152)                                      & \textbf{82.4}          & \textbf{96.6}           & \textbf{67.4}          & \textbf{91.7}           & \textbf{52.8}          & \textbf{83.8}           & \textbf{40.6}          & \textbf{74.0}           & \textbf{29.8}          & \textbf{39.6}           & \textbf{59.8}          & \textbf{75.3}           & \textbf{133.3}         & \textbf{135.4}  \\ \hline
    \end{tabular}
    }
    \caption{COCO online leaderboard of published state-of-the art image captioning models. The backbone is listed in brackets.}
    \label{table2}
    \vspace{-0.3cm}
\end{table*}

\subsection{Objectives}
Given ground truth sequence $y^*_{1:T}$ and a captioning model with parameters $\theta$.
We optimize the following cross-entropy (XE) loss:
\begin{equation}
    L_{XE}=-\sum^T_{t=1}\log(p_\theta(y_t^*|y^*_{1:t-1})).
\end{equation}
Then we continually optimize the non-differentiable CIDER-D score by Self-Critical Sequence Training (Rennie et al. 2017) (SCST) following (Cornia et al. 2020):
\begin{equation}
    \nabla_\theta L_{RL}(\theta)=-\frac{1}{k}\sum_{i=1}^k(r(y^i_{1:T})-b)\nabla_\theta\log p_\theta(y^i_{1:T}),
\end{equation}
where $k$ is the beam size, $r$ is the CIDEr-D score function, and $b=(\sum_i r(y^i_{1:T}))/k$ is the baseline.
\section{Experiments}
\subsection{Datasets}
We conduct our experiments on the benchmark image captioning dataset COCO \cite{lin2014microsoft}.
The dataset contains 123,287 images, each annotated with 5 different captions.
For offline evaluation, we follow the widely adopted Karpathy split \cite{karpathy2015deep}, where 113,287, 5,000, 5,000 images are used for training, validation, and testing respectively.
We also upload generated captions of COCO official testing set for online evaluation.
\subsection{Experimental settings}
To extract visual features, we use the pre-trained Faster-RCNN \cite{ren2015faster} provided by \cite{jiang2020defense}, that uses delated stride-1 $C_5$ backbone and $1\times1$ RoIPool with two FC layers as the detection head to train Faster R-CNN on the VG dataset.
In the feature extraction stage, it removes the delation and uses a normal $C_5$ layer to extract grid features.
For grid features, we leverage their grid features and average-pool them to $7\times 7$ grid size.
For region features, we use the same model to extract 2048-d features after the first FC-layer of the detection head.

In our implementation, we set $d_{model}$ to 512 and the number of heads to 8.
The number of layers for both encoder and decoder is set to 3.
In the XE pre-training stage, we warm up our model for 4 epochs with the learning rate linearly increased to $1\times 10^{-4}$.
Then we set the learning rate to $1\times 10^{-4}$ between $5\sim 10$ epoches, $2\times 10^{-6}$ between $11\sim12$ epoches, $4\times 10^{-7}$ afterwards.
The batch size is set to 50.
After the 18-epoch XE pre-training stage, we start to optimize our model with CIDEr reward with $5\times 10^{-6}$ learning rate and 100 batch size.
We use Adam optimizer in both stages and the beam size is set to 5. 
Following the standard evaluation criterion, we utilize BLEU@N \cite{papineni2002bleu}, METEOR \cite{banerjee2005meteor}, ROUGE-L \cite{lin2004rouge}, CIDEr \cite{vedantam2015cider}, and SPICE \cite{anderson2016spice} to evaluate our model.

\begin{figure}
    \centering
    \includegraphics[width=0.45\textwidth]{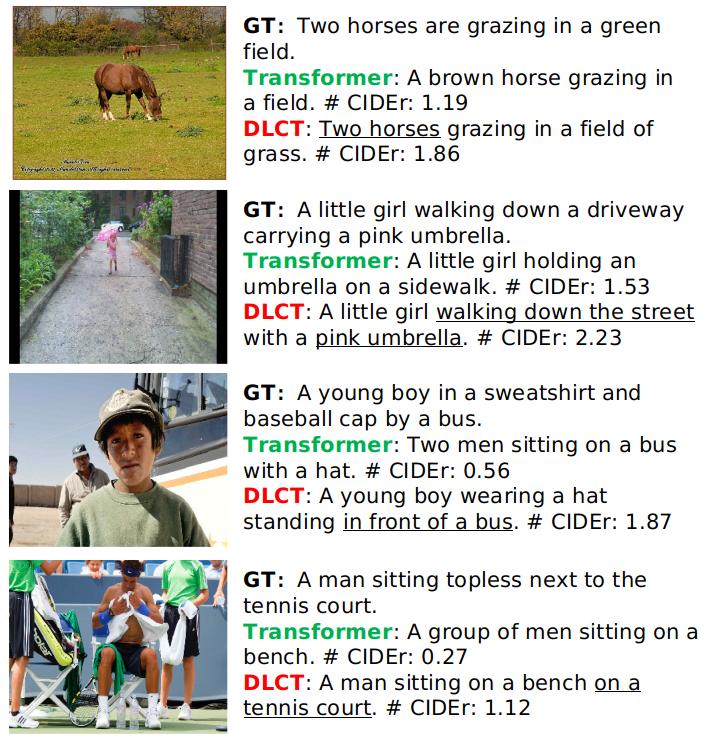}
    \caption{Examples of image captioning results by standard Transformer and our proposed DLCT with ground truth sentences and the corresponding CIDEr scores.
    Generally, our method can generate more accurate and descriptive captions.
    }
    \label{fig4}       
\end{figure}

\subsection{Performance Comparison}
\subsubsection{Offline Evaluation}
Table \ref{table1} summarizes the performance of the state-of-the-art models and our approach on the offline test split.
We also report the results of ensembled models for a comprehensive comparison.
The compared models include:
SCST \cite{rennie2017self},
Up-Down \cite{anderson2018bottom},
HAN \cite{wang2019hierarchical},
GCN-LSTM \cite{yao2018exploring},
SGAE \cite{yang2019auto},
ORT \cite{herdade2019image},
SRT \cite{Wang2020ShowRA},
AoA \cite{huang2019attention},
HIP \cite{Yao2019HierarchyPF},
M2 \cite{cornia2020meshed} and X-Transformer \cite{pan2020x}.

As shown in Table \ref{table1}, our single model consistently exhibits better performance than the others.
Our DLCT surpasses all the other models in terms of BLEU-1, BLEU-4, CIDEr while being comparable on Meteor and Rouge with the strongest competitor X-Transformer.
In sum, our DLCT outperforms X-Transformer in most of the metrics and performs slightly worse in SPICE.
The CIDEr score of our DLCT reaches 133.8\%, which advances X-Transformer by 1\%.
The boost of performance demonstrates the advantages of our DLCT which uses the complementary appearance and geometry features of regions and grids,
and models intra- and inter-level for detailed and comprehensive visual representations.
Our ensembled model achieves the best results in BLEU-1, BLEU-4, Rouge and a particularly high score in CIDEr.
Our Meteor score is comparable with the best model as well while the SPICE score is slightly worse.
For a fair comparison, we also run M2 \cite{cornia2020meshed} based on our features.
The results show that our DLCT still outperforms M2 in all metrics.

\begin{figure*}
    \centering
    \includegraphics[width=0.95\textwidth]{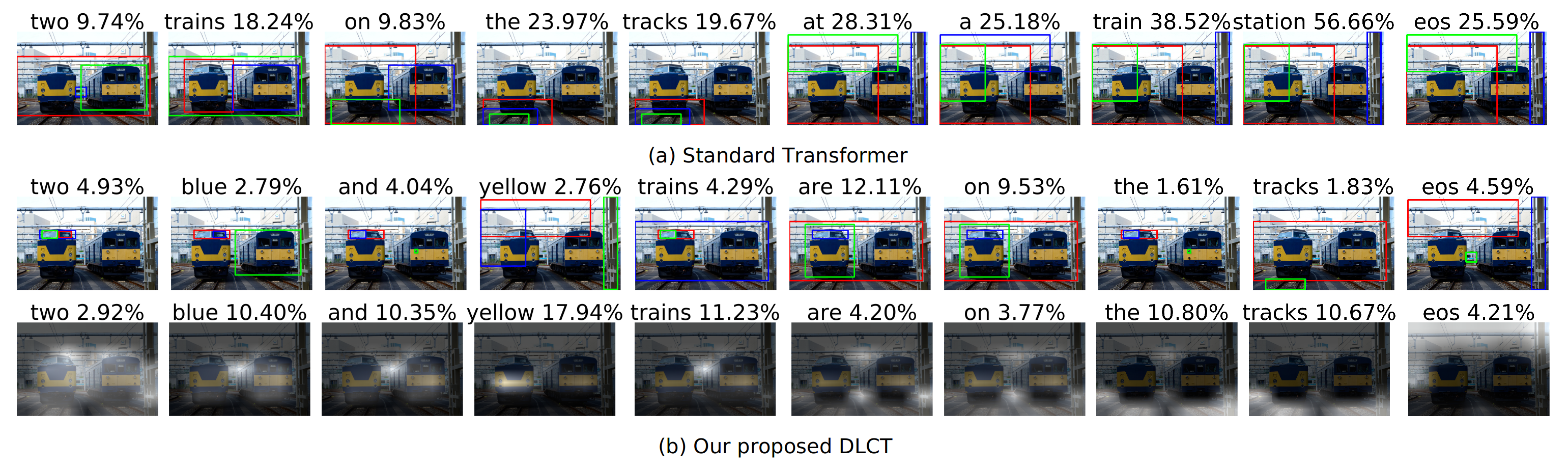}
    \caption{Attention visualization of region-based Transformer (a) and our DLCT (b).
For each word, we show top-3 attended regions (red, blue, green respectively) and the attention heatmap on grids (only available in DLCT) with the highest attention weight in the title.
Both Transformer and our DLCT can attend to corresponding regions when generating words. When generating words like ``yellow'' and ``tracks'', our DLCT can attend to corresponding grids with detailed and contextual information.
}
    \label{fig5}       
\end{figure*}

\subsubsection{Online Evaluation}
We submit the generated captions on the official testing set to the online testing server and report the results in Table \ref{table2},
which shows the performance leaderboard with 5 reference captions (c5) and 40 reference captions (c40).
For online evaluation, we ensemble 4 models and adopt two different backbones: ResNeXt-101 and ResNeXt-152 \cite{Xie2017AggregatedRT}.
Compared to all the other state-of-the-arts, our model with ResNeXt-152 achieves the best performance in all metrics.
Notably, our model with the ResNeXt-101 can achieve comparable performance to X-Transformer with SENet-154 \cite{Hu2020SqueezeandExcitationN}.

\subsection{Ablation Study}
We conduct several ablative studies to quantify the contribution of each design in our model.

\subsubsection{Features}
To better understand the effect of our features, we conduct several experiments on our features using Standard Transformer as shown in Table \ref{table3}. 
As we can see, the results of every single feature and concatenation of both features are trivial and our approach with both features can achieve much better results.

\subsubsection{CRA}
To better demonstrate the effectiveness of CRA, we conduct several ablative experiments as shown in Table \ref{table4}.
CRA can improve the performance of both the model with grid feature and the model with region feature.
And it can also improve the performance by cooperating with our LCCA, which boosts the CIDEr-D score from 133.0\% to 133.8\%.
By integrating absolute and relative location information, the captioning model can better understand the appearance features and the relationships among them.

\begin{table}[t]
	\centering
	\small
	\resizebox{0.45\textwidth}{!}{ 
    \begin{tabular}{l|llllll}
        \hline
                                      & \multicolumn{1}{c}{B-1} & \multicolumn{1}{c}{B-4} & \multicolumn{1}{c}{M} & \multicolumn{1}{c}{R} & \multicolumn{1}{c}{C} & \multicolumn{1}{c}{S} \\ \hline
        Grid (G)   & 81.2                    & 39.0                    & 29.0                  & 58.6                  & 131.2                 & 22.4                  \\
        Region (R)  & 80.1                    & 39.0                    & 28.9                  & 58.6                  & 130.1                 & 22.4                  \\ 
        G + R  & 80.9                    & 38.9                    & 29.2                  & 58.6                  & 131.6                 & 22.7                  \\ \hline
        DLCT (G+R)            & \textbf{81.4}           & \textbf{39.8}           & \textbf{29.5}         & \textbf{59.1}         & \textbf{133.8}        & \textbf{23.0}     \\ \hline    
        \end{tabular}
	}
    \caption{Performance comparison of different feature settings.}\smallskip
    \label{table3}
\end{table}

\begin{table}[t]
	\centering
	\small
	\resizebox{0.45\textwidth}{!}{ 
    \begin{tabular}{c|l|c|c|c|c|c|c}
        \multicolumn{1}{l|}{Feature} & Model           & B-1           & B-4           & M             & R             & C              & S             \\ \hline
        \multirow{2}{*}{G}           & Transformer     & 81.2          & 39.0          & 29.0          & 58.6          & 131.2          & 22.4          \\
                                     & Transformer+PE & 81.2          & 39.0          & 29.2          & 58.9          & 131.7          & 22.6          \\ 
                                     & Transformer+CRA & 81.1          & 39.3          & 29.4          & 58.9          & 132.5          & 22.9          \\ \hline
        \multirow{2}{*}{R}           & Transformer     & 80.1          & 39.0          & 28.9          & 58.6          & 130.1          & 22.4          \\
                                     & Transformer+PE & 80.6          & 38.3          & 29.0          & 58.4          & 129.7          & 22.5          \\ 
                                     & Transformer+CRA & 80.9          & 39.0          & 29.2          & 58.6          & 131.0          & 22.5          \\ \hline
        \multirow{2}{*}{G+R}         & DLCT w/o CRA    & 81.0          & 39.3          & 29.3          & 58.8          & 133.0          & \textbf{23.0} \\
                                     & DLCT            & \textbf{81.4} & \textbf{39.8} & \textbf{29.5} & \textbf{59.1} & \textbf{133.8} & \textbf{23.0} \\ \hline
        \end{tabular}
	}
    \caption{Performance with / without CRA for grids(G) and regions(R). PE represents traditional positional encoding method which directly adds positional encoding to inputs.}\smallskip
    \label{table4}
    \vspace{-0.4cm}
\end{table}

\begin{table}[t]
	\centering
	\small
	\resizebox{0.45\textwidth}{!}{ 
    \begin{tabular}{l|llllll}
        \hline
                       & \multicolumn{1}{c}{B-1}  & \multicolumn{1}{c}{B-4}  & \multicolumn{1}{c}{M}    & \multicolumn{1}{c}{R}    & \multicolumn{1}{c}{C}     & \multicolumn{1}{c}{S}    \\ \hline
        DLCT w/o LCCA  & 81.2                     & 39.2                     & 29.2                     & 58.6                     & 132.6                     & 22.8                     \\
        LCCA + CBG & \multicolumn{1}{r}{80.8} & \multicolumn{1}{r}{38.7} & \multicolumn{1}{r}{29.0} & \multicolumn{1}{r}{58.7} & \multicolumn{1}{r}{130.8} & \multicolumn{1}{r}{22.7} \\ \hline
        DLCT           & \textbf{81.4}            & \textbf{39.8}            & \textbf{29.5}            & \textbf{59.1}            & \textbf{133.8}            & \textbf{23.0}        \\ \hline
        \end{tabular}
    }
    \caption{Performance with / without LCCA, where CBG means the complete bipartite graph.}
    \label{table5}
    \vspace{-0.3cm}
\end{table}

\subsubsection{LCCA}
We also conduct several experiments to demonstrate the effectiveness of our LCCA, which are shown in Table \ref{table5}.
Two alternatives are considered:
one is our DLCT without LCCA,
and the other is LCCA with a complete bipartite graph (CBG) in which cross attention is applied between all grid nodes and region nodes.
They both show worse performance than LCCA, which demonstrate the superiority of our LCCA.
Note that DLCT with CBG is even worse than standard Transformer with grid feature inputs,
which shows the damage of semantic noises introduced by coarsely modeling relationships between regions and grids.

\subsection{Qualitative results and visualization}
Fig. \ref{fig4} illustrates several example image captions generated by Transformer and DLCT.
As indicated by these examples, generally, our DLCT can grab detailed and contextual information to generate more accurate and descriptive captions.

In order to better qualitatively evaluate the encoded visual representations, we visualize the contribution of each visual feature to the model output in Fig \ref{fig5}.
Technically, we average attention weights of 8 heads in the last Enc-Dec Multi-head Attention Layer.
We can see that both Transformer and DLCT are able to attend to the corresponding regions when generating words.
In addition, our DLCT can attend to corresponding grids when it generates the word ``blue" and ``yellow".
When generating the word ``tracks", the attention heatmap on grids provides a more fine-grained semantic segmentation of tracks,
which demonstrates the advantages of our DLCT.

\section{Conclusion}
In this paper, we proposed a Dual-Level Collaborative Transformer to achieve the complementarity of region and grid features for image captioning.
Our model integrates appearance and geometry features of regions and grids by applying intra-level fusion via Comprehensive Relation Attention (CRA) and Dual-Way Self Attention (DWSA).
We also proposed a geometric alignment graph to apply Locality-Constrained Cross Attention (LCCA) which helps reinforce two kinds of features effectively and  address the issue of semantic noises aroused by the direct fusion of two sources of features.
Extensive results demonstrate the superiority of our approach that achieves a new state-of-the-art on both offline and online test splits.
In our feature work,  we plan to extend the proposed collaborative features to other multi-media areas which requires detailed and contextual information.

\section{ Acknowledgments}
This work is supported by the National Science Fund for Distinguished Young (No.62025603), the National Natural Science Foundation of China (No.U1705262, No. 62072386, No. 62072387, No. 62072389, No. 62002305,  
No.61772443, No.61802324 and No.61702136) and and Guangdong Basic and Applied Basic Research Foundation (No.2019B1515120049).

\bibliography{aaai2021_cite}
\end{document}